# Finding the optimal human strategy for Wordle using maximum correct letter probabilities and reinforcement learning

Benton J. Anderson[1], Jesse G. Meyer[2]

**Affiliations:**
[1]University of Wisconsin-Madison, Madison, Wisconsin, USA, bentonanderson@outlook.com
[2]Medical College of Wisconsin, Milwaukee, Wisconsin, USA, jessegmeyerlab@gmail.com

**Abstract**

Wordle is an online word puzzle game that gained viral popularity in January 2022. The goal is to guess a hidden five letter word. After each guess, the player gains information about whether the letters they guessed are present in the word, and whether they are in the correct position. Numerous blogs have suggested guessing strategies and starting word lists that improve the chance of winning. Optimized algorithms can win 100% of games within five of the six allowed trials. However, it is infeasible for human players to use these algorithms due to an inability to perfectly recall all known 5-letter words and perform complex calculations that optimize information gain. Here, we present two different methods for choosing starting words along with a framework for discovering the optimal human strategy based on reinforcement learning. Human Wordle players can use the rules we discover to optimize their chance of winning.

**INTRODUCTION**

Wordle is a popular online word puzzle game invented by Josh Wardle that has become immensely popular around January 2022 [1]. The goal of the game is to guess a hidden five letter word in no more than six guesses. There are 2,315 possible secret words and the game accepts 12,972 possible words as guesses. After every guess, the computer reveals whether each letter is absent from the word (gray), present in the word but in a different position (yellow), or present in the correct position (green). Wordle is unique in that it only allows one game to be played per day, and every player in the world plays to guess the same word each day.

With hundreds of thousands of daily players, it is common to ask: what is the optimum strategy to win Wordle? Because Wordle is forgiving in the number of guesses and the amount of information given at each step, it is relatively simple to construct a program that guesses a first optimal word, then uses the new information at each step to maximize information and win 100% of games. With the large variety of possible English words, and the requirement that guesses be valid 5-letter English words, it can be difficult as a human without perfect recall of all words to discern a sufficiently informative next word to guess. We seek to define the optimum human strategy question in two ways: one from a human perspective with an easy heuristic that gives a list of words to guess at the start of every game, and another from a machine perspective that tests the set of possible actions at various game states.

While numerous machine algorithms can arrive at an optimal solution, the ideal Wordle strategy for a human is highly debatable, as shown by the numerous blog posts describing different optimal first words to guess. One article claims the best word is "LATER" [2], but other analyses claim "SOARE", "RAISE", "ROATE", or "SLATE", among others [3]. There are countless blog posts and news articles that claim to solve Wordle, but we are unaware of any prior reports that used machine learning.

Previous studies have shown that reinforcement learning can effectively create policies that win games that cannot be completely solved mathematically. Notably, reinforcement learning was used in AlphaGO to achieve superhuman performance at the game "Go" in 2016 [4]. Reinforcement learning has even been used for multiple simple Atari computer games [5] and more complex real-time strategy games such as DotA 2 [6]. Despite the ability of computers to master Wordle using complete knowledge of the vocabulary, to optimize the use of various human strategies for playing Wordle at different game states, we used reinforcement learning. Specifically, we used model-free q-learning [7]. Reinforcement learning is useful for this task because it allows us to test all possible strategies at different game states for thousands of game instances. Q-learning returns an optimal policy for each of the game states, which can guide human players to be more effective at the game Wordle.

**METHODS**

**Data and Code**
The word database containing 12,972 total words was downloaded from the Wordle website (https://www.powerlanguage.co.uk/wordle/), with the first 2,315 words used as possible secret words, and all words allowed as possible guesses. The complete list of words is available on github along with the code https://github.com/benton-anderson/wordle-opt.

**Wordle implementation**
For testing, Wordle was implemented as a class in Python. Wordle is a game where a secret five letter word is chosen and the player must guess the word within six tries. After each guess, the player gains information. If the letter is gray, that letter is not present in the word. If the letter turns green, that letter is present in the word at that exact position. If the letter turns yellow, that letter is present in the word, but at a different position. The python class reproduced these features using attributes that track the word list and the current state of the gameplay including the number of green and yellow letters. The class method 'try_word' takes as input a guess word, compares it with the secret word, updates the state of the game, and returns the result.

**Calculating green letter probability**
As a metric for determining probability of a letter appearing as a green in a given letter position, a bag-of-letters approach (similar to bag-of-words used in natural language processing) was calculated on the population of 2,315 words. The probability of each letter returning a green is given in **Figure 1A**.

**Calculating word log-likelihood based on green letter probability distribution**
With the bag-of-letters used as 5 separate categorical distributions of 26 possible outcomes ( corresponding to the letters A through Z), the log-likelihood (*LL*) of each word arising from a random sampling of these distributions is given by equation (1)

$$(1) \quad LL = \sum_{i=1}^{5} \log(Pr(X_i = \alpha))$$

Where $\alpha$ is the letter derived from each of the 5 categorical distributions at position *i*. Note that for certain letters in some positions, the probability is 0, therefore Laplace estimates (*n=1*) were added to each probability to avoid -inf logarithms.

**Calculating optimum Total Green Letter Probability (TGLP) for sequences of N words.**
For each word in the database, the green letter probability is calculated and the words are ranked from highest to lowest green letter probability. Iterating through all 12,972 words, a greedy search algorithm constructs sequences of words that aim to maximize green letter probability for the next word in the sequence, given that a minimum of letters are re-used from the previous word guesses. In this way it sums the green letter probabilities of each word and generates a total green letter probability (TGLP) using equation (2)

$$(2) \quad TGLP = \sum_{j=1}^{N} \sum_{i=1}^{5} Pr(X_i = \alpha)$$

Where $\alpha$ is the letter derived from each of the 5 categorical distributions at position *i*, and *j* is the green letter probability for each of N words in the word guess sequence.

**Reinforcement Learning**
Reinforcement learning was carried out in python using the gym package [8]. Three classes were the game "Wordle", the agent "Player", and the "Environment". Q learning was used to allow the agent to find an optimal policy for playing Wordle based on rewards. Rewards were: yellow=2, green=5, win=25, lose=-15. The number of games was defined as 1,000 or 10,000, epsilon=0.3 or 0.02, alpha=0.02, and discount_factor=0.05. State representations were compared using either a tuple of the number of greens and yellows in the last guess, or the number of greens and yellows ever guessed for each position. The notebook enabling anyone to recreate this reinforcement learning is available on the github repository.

The agent was allowed to make one of five types of guesses each round of the game: (1) random, (2) word likelihoods derived from letter probabilities, (3) green letter probability maximization, (4) smart, or (5) exclude. The random guess randomly picks one of the words in the word list. The letter probability list ('probs1') guesses one of the words from the defined list of four starting words in order: BOWNE, SLATY, PRICK, FAUGH and MEVED. The green letter maximization word list ('probs2') guesses one of the following words in order: LOOIE, SAURY, CHANT, BIDED and PRIMP. If the reinforcement learning algorithm picks either probs1 or probs2 for a sixth time within one game, these guess methods return a random word from the remaining words. "Smart" guess uses all the available information to guess, including the number and positions of yellow and green letters, and excludes words that contain letters in the absent list. "Exclude" only excludes words from the word list that contain letters that are known to be absent from the list, which would be the gray tiles in the game.

Policy heatmaps were generated by averaging three policy dictionaries from three repeated reinforcement learning experiments with 10,000 iterations each. The value for each state (row) was normalized to a range of 0 to 1.

## RESULTS

### Dataset characters
The letter probabilities shown in **Figure 1A** give the categorical distribution of the population of letters found in each letter position, where each column sums to 1. In the first column, the letter 'S' has the highest probability, indicating that 'S' is the most common starting letter for 5-letter words. Similarly, the 5th position shows that 'E' is the most common ending letter. This distribution was calculated only on words found in the 2,315 word database as opposed to the full 12,972 possible guesses in Wordle. We chose to focus only on the possible winning answers for this analysis because our goal is to find the best guesses for human players. While this may be seen as biasing the training data toward only those words that can win, we argue that with the entire population of possible correct words known beforehand, this choice creates a foundation for our method that makes winning as a human most likely, which is the fundamental goal of this work.

### Optimal word patterns
This method aims to optimize a Wordle strategy for a human player, rather than a machine. While machines can make perfect use of information from grays and yellows in choosing their next move, they represent only a slight boost in certainty for a human. Whereas with a green letter, a human can better construct their next move with the certainty that a green provides.

With this rationale for prioritizing greens in mind, the probabilities for each letter in all five positions given in **Figure 1A** represents the probability of a green arising when guessing a word. There are multiple methods to best make use of this information. The first calculates word likelihoods from letter probabilities as the value to maximize. Here, likelihoods represent the probability that this specific sequence of letters would arise from the **Figure 1A** distribution. The second method sums each letter's probability to derive a total green letter probability that can be maximized. While each method begins from the same data distribution, the information given in each is somewhat correlated, but not perfectly linear (**Figure 1B**) suggesting that each method will have its own benefits.

While we prioritize greens over grays and yellows, we do not fully discount their utility, so in our search for the ideal 3-word sequence for maximum information, we penalize repeating letters by looking first for the word that maximizes the addition to TGLP without any repeated letters. If no such word exists because the list has been exhausted of words that would not give a repeat letter from the prior words, then we look for words with only one repeat.

Though it is ideal to avoid repeat letters in the word sequence to maximize information gained, it is notable that words with repeat letters do not inherently correlate with a lower TGLP score. Comparing distributions of TGLP between words (**Figure 1C**) with 5, 4, 3 and 2 unique letters in the word, we see that the vast majority have greater than 2 unique letters, and of those, the TGLP distributions are approximately the same. This suggests that when searching for optimal word sequences, the penalty for repeating letters already used is not a significant loss in TGLP, as long as the repeat letters do not reoccur in the same position.

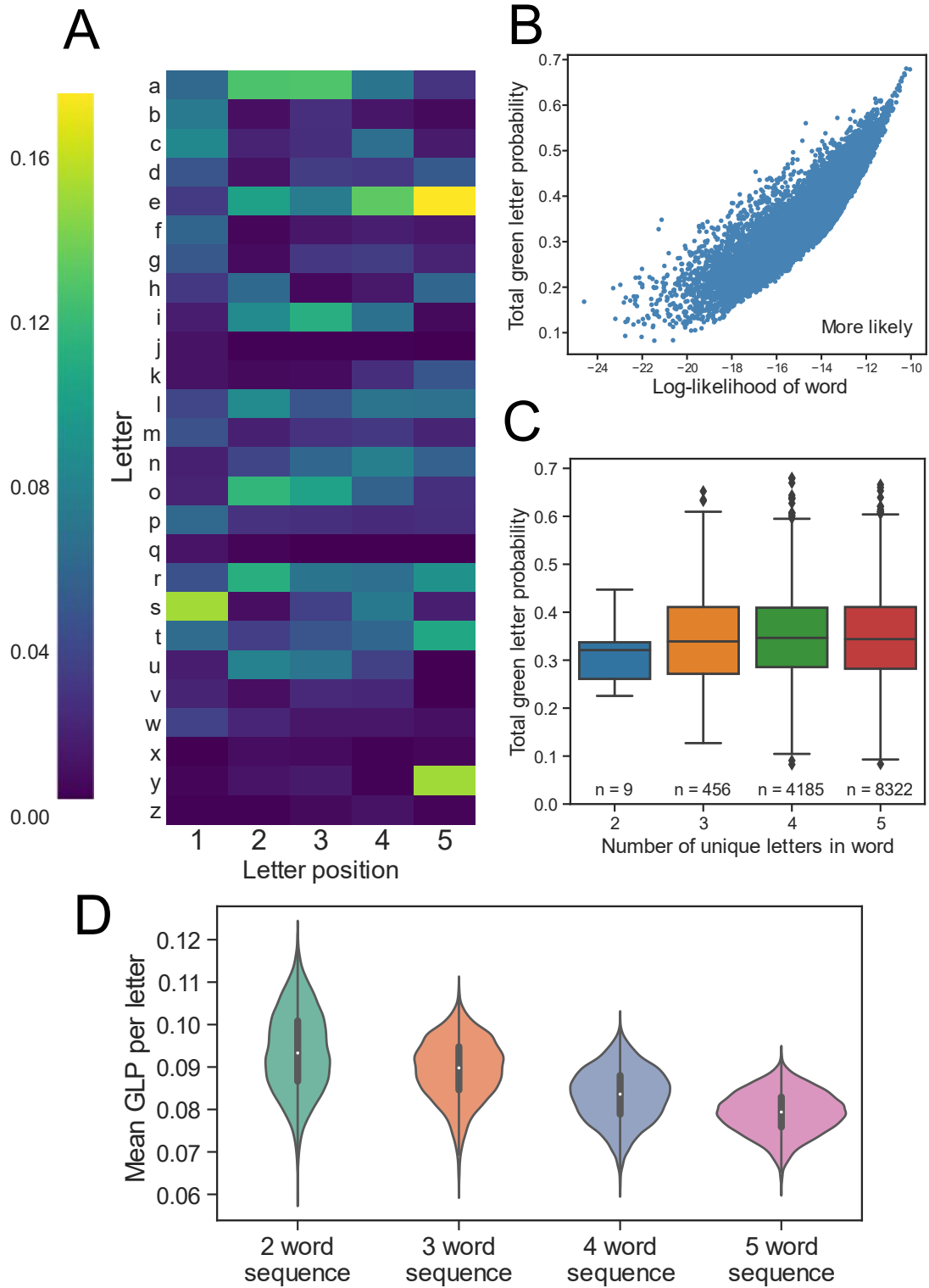

**Figure 1: Picking words based on letter probabilities A**. Heatmap showing the frequencies of each letter occurring at each position in a five-letter word. The values in each column sum to equal 1. **B.** Total green letter probability (y-axis) versus log-likelihood of all 12,972 words. **C.** Relationship between total green letter probability and the number of unique letters in a word. **D.** Mean contribution to total green letter probability per letter for each sequence after search algorithm across all words, separated by number of words in sequence. The highest-valued result at the top of each distribution represents the optimal N-word sequence as given in Table 1.

For generating optimum word lists, we constructed a greedy search algorithm that aimed to maximize the total score of a sequence of words as described in methods. The Algorithm generated a sequence of words that maximized each score. As shown in **Figure 1D**, the average contribution to TGLP per letter used decreases as the word sequence grows longer, suggesting that the search algorithm cannot find later words that are as efficient as the first words based on our constraints. Because the algorithm is greedy at each step, we can calculate a mean TGLP contribution per unique letter, and get a metric of efficiency of the TGLP method. In this way we come to 5 optimum word lists, depending on if the player wants to guess a 1-, 2-, 3-, 4-, or 5-word sequence in their game.

**Table 1: Optimal guess sequences via search with TGLP score**

| Sequence length | | | | | |
|---|---|---|---|---|---|
| 1-word | SAREE | | | | |
| 2-word | SEINE | COALY | | | |
| 3-word | POUPT | SAINE | CLARY | | |
| 4-word | AURAE | SOILY | CHANT | FADED | |
| 5-word | LOOIE | SAURY | CHANT | BIDED | PRIMP |

A similar search using the word log-likelihood, maximizing summed log-likelihoods of words reveals a different list (**Table 2**):

**Table 2: Optimal guess sequences via search with log-likelihood score**

| Sequence length | | | | | |
|---|---|---|---|---|---|
| 1-word | SOOEY | | | | |
| 2-word | SOILY | CRANE | | | |
| 3-word | SHIED | COALY | BRUNT | | |
| 4-word | PLATE | SHINY | BROCK | FUMED | |
| 5-word | BOWNE | SLATY | PRICK | FAUGH | MEVED |

**Reinforcement Learning**
Given the range of strategies players can choose, and how those strategies interact with the information available to the player, we used reinforcement learning with 10,000 iterations to test different combinations of actions at different game states. This allowed us to automatically determine the optimal game policy. Q-learning was used in python to emulate the Wordle game, a player agent, and an environment that allowed the game and player to interact (**Figure 2A**). The system allowed the player to make one of five actions, which were guesses based on:
1. A random draw from the remaining words ('random')
2. The four words from word likelihoods derived from letter probabilities ('prob1')

3. The four words from total green letter probability maximization ('prob2')
4. Knowledge of the green letters, yellow letters, and absent letters ('smart')
5. Excluding words that have letters known to be absent ('absent').

After each round of the game, the action was rewarded based on how many yellow letters (+2 points) and green letters (+5 points). The framework also awarded a win with +25 points, or a loss with -15 points. An example of the reward distribution across 10,000 trials is shown in **Figure 2B**.

The current state of the game could be encoded in multiple ways. Since the major determinants influencing real human players are the counts of green and yellow letters, a tuple giving the number of green and yellow hits were used to represent the state. Using the count of green and yellow letters from the last guess ('last') was compared with using the total count of green and yellow letters ('best') for three repeated reinforcement learning trials. Comparing the win rate of these two schemes over the first 1,000 iterations showed that using the result from the last guess was significantly better than using the best count (**Figure 2C**). The average number of wins for three trials of 10,000 games using the last result as state was 4,970 with a standard deviation of 83, whereas the average of the trials using the best result as state was 4,664 with a standard deviation of 110. This difference is significant according to a two-sided independent t-test assuming unequal variance with p-value=0.04.

A total win rate of 64.8% was achieved after 10,000 trials using the best parameters, which were epsilon=0.02, alpha=0.02, and discount_factor=0.05. Most games lasted all six rounds (**Figure 2D**). The player agent won once on the first try, 52 times on the second try, or 415 times on the third try.

The key result from the reinforcement learning comes from analysis of the learned policy (**Figure 2E, 2F**). This revealed an optimal strategy to use under various game situations based on the five possible actions of the agent. When the last guess yields no green or yellow hits, the player gets the most rewards by following the list of rational word choices that optimize the chance of green letters ('probs1' or 'probs2', top row). As soon as the model has any number of yellow hits except five (rows 2-6), the optimal choice for the agent is the guess based only on excluding the letters that are known to be absent. When the model has at least one green hit (all other rows), the best action is to use all available information including the yellow and green letters, and to exclude the letters that are known to be absent.

The optimal policy depended on the value of epsilon used for reinforcement learning, which determines the balance between exploiting learned actions and exploring new actions. This was most apparent when the agent has no yellow and no green letters in the current state. When epsilon was set to 0.5 and exploration was high (**Figure 3E**), the model favors the second word list 'probs2' based on TGLP. When epsilon was set to 0.02 and exploitation was high (**Figure 3F**), the system favors the first word list 'probs1' based on the word likelihood.

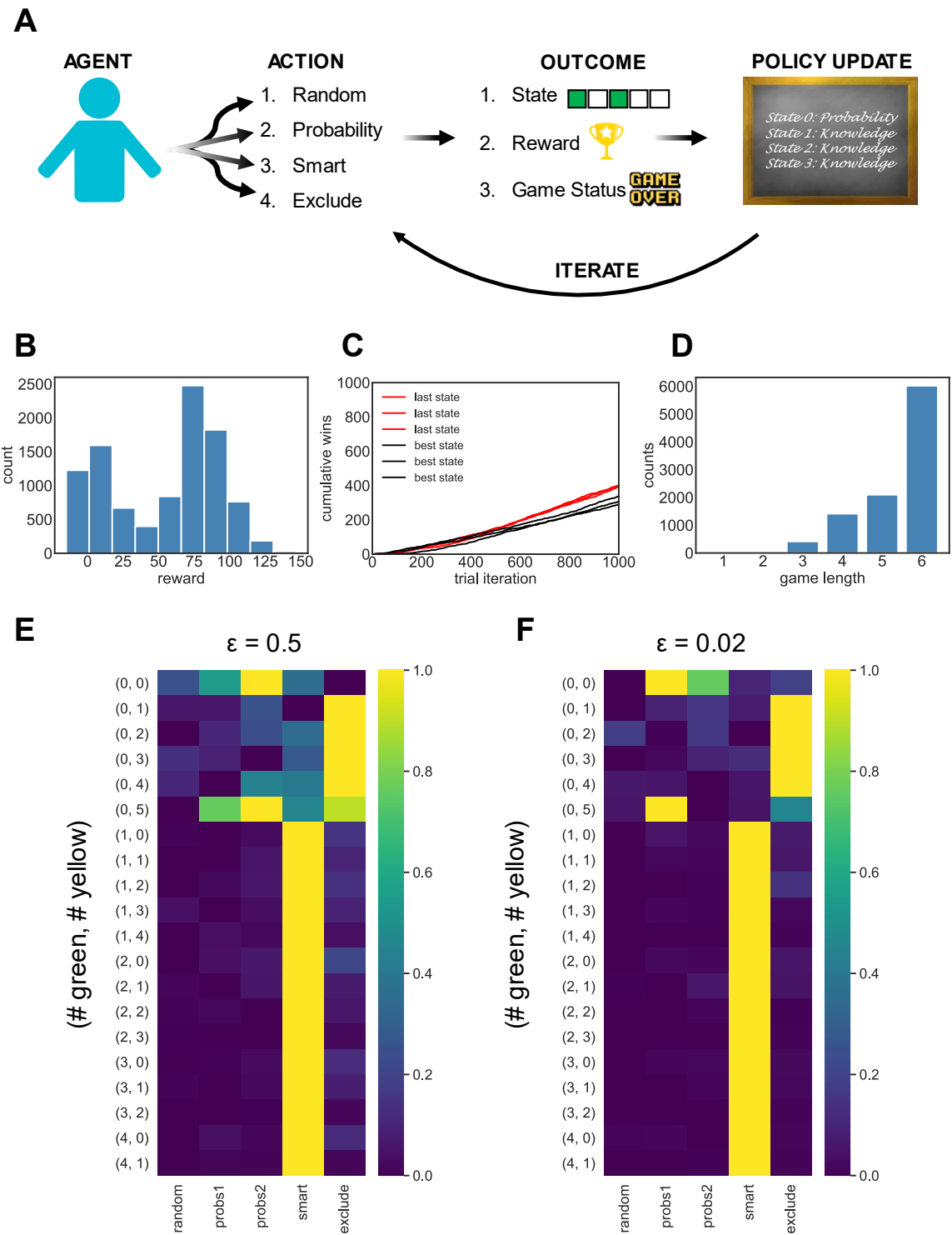

**Figure 2: Reinforcement learning optimizes rules for humans playing wordle. A.** Scheme showing the concept of reinforcement learning for this task of Wordle. **B.** Distribution of game rewards at the end after 10,000 trials. Given the large size of the reward for winning at 50, the distribution is bimodal. **C.** Comparison of 'best' versus 'last' game state representations for three repeated learning experiments with each setting. 'Best' defines the state as the highest total counts of green and yellows and 'last' defines the state as the counts of green and yellow in the last guess. **D.** Distribution of game lengths over 10,000 trials of the game using 'last' as the game state. Most games lasted for all six trials. **E, F.** Optimal policy of overview from an average of the policy learned from 10,000 trials repeated three times using epsilon of 0.5 (E, more exploration) versus 0.02 (F, more exploitation). The row labels give the state of as defined by (# of green letters, # of yellow letters). Each row is normalized so that the highest value is set to 1 and the lowest value is set to zero.

# DISCUSSION

Although other work has shown how Wordle can be solved with a 100% success rate by maximizing the information gained from each guess, this strategy is not possible for a human player [9]. Our word picking strategies combined with a reinforcement learning approach enabled optimization of actions at different game states. This allowed us to discover an optimized strategy for a human playing Wordle. That strategy is:

1. Guess from the list of words that are most likely to hit a green letter until the player has at least one green or yellow letter. Based on the two tested word lists, the following word list is best: BOWNE, SLATY, PRICK, FAUGH and MEVED.
2. If the player gets at least one yellow letter, they should guess words that can exclude more letters, while ignoring whether those words contain yellow letters.
3. If the player gets at least one green letter, they should use all available information about letters that are and are not present to limit the word list for their next guess.

This strategy was surprising because it suggests that even when the player has information about yellow letters, it is better to try to exclude additional letters. This strategy is similar to some other machine strategies that aim to increase the information content.

An important caveat is that although we report one optimal strategy here, this is only optimal within the set of action types we tested. There is much room to improve the possible set of actions. The reward system can also be optimized.

In conclusion, this work provides a starting point for human-centric Wordle strategies that gives two new ways to prioritize word lists. We demonstrated that q-learning can optimize a set of potential Wordle actions for specific game states. We expect the rules we discover can be widely used by human players to increase their odds of winning. We also expect that the frameworks we provide will enable even better human policies to be developed for Wordle in the future.

# ACKNOWLEDGEMENTS
This work was brought to you by the letters BJA and JGM, and the number 2,315.